\documentclass[11pt,a4paper]{article}
\usepackage{ifthen}
\usepackage[hyperref]{eacl2021}
\usepackage{times}
\usepackage{latexsym}
\usepackage{tikz}
\usepackage{ctable}
\usepackage{bm}
\usepackage[textwidth=2cm, textsize=small]{todonotes}
\usepackage{ulem}
\usepackage[acronym]{glossaries}

\newacronym{crf}{CRF}{Conditional Random Field}
\newacronym{rnn}{RNN}{Recurrent Neural Networks}
\newacronym{lstm}{LSTM}{Long Short-Term Memory}
\newacronym{blstm}{BiLSTM}{Bidirectional Long Short-Term Memory}
\newacronym{bgru}{BiGRU}{Bidirectional Gated Recurrent Unit}
\newacronym{cnn}{CNN}{Convolutional Neural Network}
\newacronym{asf}{ASF}{Apache Software Foundation}
\newacronym{svm}{SVM}{Support-vector Machine}
\newacronym{nlp}{NLP}{Natural Language Processing}

\usepackage{microtype}
\usepackage{float}
\usepackage{soul}

\aclfinalcopy % Uncomment this line for the final submission
\title{Multilingual Email Zoning} % with cross lingual embeddings

\author{Bruno Jardim  \\
  Cleverly, Lisbon, Portugal \\
  NOVA-IMS, Lisbon, Portugal \\
  \texttt{bjardim@novaims.unl.pt} \\
  \And
  Ricardo Rei \\
  NOVA-IMS, Lisbon, Portugal \\
  Unbabel, Lisbon, Portugal \\
  \texttt{rrei@novaims.unl.pt} \\
  \AND
  Mariana S. C. Almeida \\
  Cleverly, Lisbon, Portugal \\
  \texttt{mariana.almeida@cleverly.ai} \\
}

\date{}

\begin{document}
\maketitle

\begin{abstract}

The segmentation of emails into functional zones (also dubbed \textbf{email zoning}) is a relevant preprocessing step for most NLP tasks that deal with emails.
However, despite the multilingual character of emails and their applications, previous literature regarding email zoning corpora and systems was developed essentially for English.

In this paper, we analyse the existing email zoning corpora and propose %Cleverly zoning corpus,
a new multilingual benchmark composed of 625 emails in Portuguese, Spanish and French. Moreover, we introduce {\sc{Okapi}}, the first multilingual email
segmentation model based on a language agnostic sentence encoder. Besides generalizing well for unseen languages,
our model is competitive with current English benchmarks, and reached new state-of-the-art performances for domain adaptation tasks in English.

\end{abstract}

\section{Introduction}

Worldwide, email % has been 
is a predominant means of social and business communication. Its importance has attracted %various
studies in %the 
areas of %social sciences, 
Machine Learning (ML) and Natural Language Processing (NLP),
 impacting a wide range of applications, from spam filtering \cite{QaroushKW12} to 
network analysis \cite{Christidis2019}.

\begin{figure}[h!]
  \centering
  \includegraphics[width=0.4\textwidth]{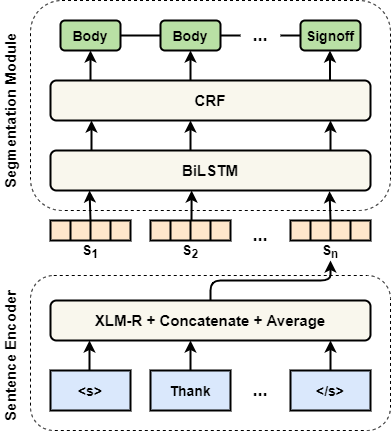}
  \caption{{\sc{Okapi}} is composed of two building blocks:  
  1) a multilingual sentence encoder (XLM-RoBERTa) 
  to derive sentence embeddings; and 2) a segmentation module that uses a BiLSTM with a CRF on top to classify each sentence into an email zone.}
  \label{fig:Okapi-Model}
\end{figure}

The email body is commonly perceived as unstructured textual data with multiple possible formats. %and difficult to process, 
However, it is possible to discern a level of formal organization in the way most emails are formed. Different functional parts can be identified such as greetings, signatures, quoted content, legal disclaimers, etc. %and  advertisement. 
The segmentation of email text into zones, also known as \textbf{email zoning} \cite{LampertDP09}, has since become a prevalent preprocessing task for a diversity of downstream applications, such as author profiling \cite{Estival07authorprofiling}, request detection \cite{lampert-etal-2010-detecting}, uncover of technical artifacts \cite{Bettenburg11}, automated template induction \cite{Proskurnia17}, email classification \cite{Kocayusufoglu19} or automated email response suggestion \cite{45189, Chen19}.

Since email communication is a worldwide phenomenon, 
all previous applications 
are in fact highly multilingual.
Despite this, email zoning literature remains
English-centric and without a standardize zone taxonomy. 
To mitigate those problems, we make the following research contributions:
\begin{enumerate}
    \item We discuss the existing zoning taxonomies and their limitations.
    \item We release Cleverly zoning corpus, the first multilingual corpus for email zoning. This corpus consists of 625 emails in 3 languages rather than English (Portuguese, Spanish and French), and encompasses 15 email zones as defined in \citep{BevendorffKPS20} %não devia ser cite?
    \item We introduce {\sc{Okapi}}, a multilingual email segmentation system built on top of XLM-RoBERTa \cite{conneau2019unsupervised} that can be easily extended to 100 languages.
\end{enumerate}

To the best of our knowledge, {\sc{Okapi}} is the first end-to-end multilingual system exploring pre-trained transformer models \cite{NIPS2017_3f5ee243} to perform email zoning. Besides having multilingual capabilities, {\sc{Okapi}} is competitive with existing approaches for English email zoning, and attained state-of-the-art performance in domain adaptation tasks for English email zoning.

The rest of the paper is organized as follows: Section \ref{sec:literature-review} presents an overview of the related literature. Section \ref{sec:Corpora} provides a comprehensive review of existing email zoning corpora, and introduces Cleverly zoning corpus, our new multilingual email zoning corpus. Section \ref{sec:okapi-model} describes the {\sc{Okapi}} model architecture. 
Section \ref{sec:results_discussion} reports and discusses the 
results achieved. Finally, Section \ref{sec:conclusion} concludes the paper. 

\section{Literature Review}
\label{sec:literature-review}

\begin{table*}[ht]
\centering
\begin{tabular}{@{}llrrl@{}}
\toprule
\multicolumn{1}{c}{Authors}                               & Source  & emails & zones & Language \\ \specialrule{1.5pt}{1pt}{1pt} 
\citep{CarvalhoC04}$^4$ & 20 Newsgroup$^5$ & 617       & 2        & English \\
\citep{Estival07authorprofiling}$^6$ & Donated$^6$  & 9,836     & 5        & English      \\
\citep{LampertDP09}$^7$                      & Enron$^8$        & 400       & 3/9      & English      \\ \hline
{\citep{RepkeK18}}$^9$ & Enron$^8$        & 800       & 2/5      & English      \\
                                                          & ASF\footnotemark[1]          & 500       & 2/5      & English      \\ \hline
{\citet{BevendorffKPS20}}\footnotemark[10] & Gmane$^3$        & 3,033     & 15       & Multilingual* \\
                                                          & Enron$^8$        & 300       & 15       & English      \\ \hline
\textbf{Ours} & Gmane$^3$ & 625   &  15    & Multilingual\\           \bottomrule
\end{tabular}
\caption{Summary of existing email zoning corpora.
*Note that, although \citet{BevendorffKPS20}'s Gmane corpus
is technically multilingual, it only has 38 non-English test emails that are spread over 13 different languages.}
\label{tab:corpus/sources/zones}
\end{table*}

\citet{Chen99} were one of the pioneers in the topic of email segmentation. Looking at linguist and geometrical patterns, their work focuses on the identification of email signature. Similarly, \citet{CarvalhoC04} developed {\sc{Jangada}}, a supervised learning system that classifies each line using a Conditional Random Field (CRF) \cite{Lafferty2001} and a sequence-aware perceptron \cite{Collins02}, %attempting to identify
that identifies signature blocks and quoted text from previous emails. \citet{Tang2005} proposed an email data cleansing system based on a Support Vector Machine (SVM) \cite{CortesSMV95} model that aimed at filtering the non-textual noisy content from emails independently of downstream text mining applications, 
based on hand-coded features.

\citet{Estival07authorprofiling} were the first to introduce a general segmentation schema for email text. Segmentation of emails is a crucial part on their work, which aims at identifying the author’s basic demographic and psychometric traits. In that work, the authors compared a range of ML algorithms together with feature selection to classify email segments into five functional parts, %(author text, signature, advertisement, quoted text, and reply lines), 
attaining improvements in the end task of auto profiling. 
Later, \citet{LampertDP09} formally defined the functional parts as email zones, describing the different segments inside email messages based on graphic, orthographic, and lexical features. 
\citet{LampertDP09} also proposed {\sc{Zebra}}, an email zoning system based on a SVM. In a posterior work towards detecting emails containing requests for action, \citet{lampert-etal-2010-detecting} used {\sc{Zebra}} to ``zone'' emails, considering only the zones that had relevant patterns to increase the accuracy of their request detection task. 

As email zoning surpassed its original purpose of signature identification and text cleansing into a more general task, \citet{RepkeK18} extended its utility to thread reconstruction. Inspired by {\sc{Zebra}} \cite{LampertDP09}, the authors proposed {\sc{Quagga}} \cite{RepkeK18}, a neural system with a Convolutional Neural Network (CNN) \cite{LeCun89} to produce sentence representations followed by a Recurrent Neural Network (RNN) \cite{Elman90}.
{\sc{Quagga}} was trained and evaluated on English emails from both the Enron \cite{Lang95} corpus and the public mail archives of the Apache Software Foundation (ASF)\footnote{\url{http://mail-archives.apache.org/mod\_mbox/}},  outperforming {\sc{Jangada}}
and {\sc{Zebra}}.

Until very recently, email zoning resorted to small samples of mailing lists or newsgroup corpus and was limited to the English language. \citet{BevendorffKPS20} were the first to crawl email at scale, extracting 153 million emails from the Gmane email-to-newsgroup gateway\footnote{\url{https://news.gmane.io/}} in different languages such as English, Spanish, French and Portuguese\footnote{\url{https://webis.de/data.html?q=Webis-Gmane-19}}. 
The authors annotated email zones for a subset of Gmane English emails and, due to the idiosyncratic characteristics of the corpora, they developed a more fine-grained zone classification schema with 15 zones. Moreover, \citet{BevendorffKPS20} introduced an email zoning system, named {\sc{Chipmunk}}, that combines a Bidirectional Gated Recurrent Unit (BiGRU) \cite{Cho14} with a CNN.
When compared to other models in the literature, {\sc{Chipmunk}} achieved better performance. 

\section{Email Zoning Corpora}
\label{sec:Corpora}

Several corpora and zoning schemes have been proposed in the literature under different contexts. This section provides an overview of the existing corpora, hoping to make it easier to develop and compare new email zoning methods in the future.

Table \ref{tab:corpus/sources/zones} compiles the information of existing email zoning corpus. To the best of our knowledge, \citet{CarvalhoC04} released the first email zoning corpus. The corpus consists of 617 emails\footnote{\label{note1}\url{http://www.cs.cmu.edu/~vitor/codeAndData.html}} from the 20 Newsgroup corpus\footnote{\url{http://qwone.com/~jason/20Newsgroups/}} identified with two zones: \textit{signature} and \textit{quotation}. Despite the usefulness of identifying those zones for email cleansing, this level of detail is still insufficient for a general email segmentation.  

\citet{Estival07authorprofiling} released a corpus of 9,836 recruited respondents donated email messages\footnote{available upon contact with the authors.} and introduced a wider annotation schema focusing on more email parts: \textit{author text}, \textit{signature}, \textit{advertisement}, \textit{quoted text}, and \textit{reply} lines. However, \citet{Estival07authorprofiling} still did not divide the email text into some other relevant zones, such as greetings, closings nor identify attachments and code lines.

\citet{LampertDP09} were arguably the first to conceptualize the email zoning task and fully define the characteristics of each identified zone, as well as dividing the authored text into different zones. They annotated 400 English emails\footnote{\url{http://zebra.thoughtlets.org/}} from the Enron email corpus database dump, identifying 3 email zones: \textit{sender}, \textit{quoted conversation} and \textit{boilerplate} zones, each containing a different set of sub-zones, within a total of 9 sub-zones. 

\citet{RepkeK18} also resorted to the Enron database\footnote{\url{http://www.cs.cmu.edu/~enron/}}, annotating a total of 800 emails\footnote{\url{https://github.com/HPI-Information-Systems/Quagga}}. Reconsidering the task as thread reconstruction, they produced a new annotation schema, considering a 2-level and a 5-level approach (the latter being a refinement of the 2-level segmentation). 
\citet{RepkeK18} also annotated 500 ASF emails$^7$ using both the 2-level and 5-level taxonomies. Their 5-level annotation schema consists of segmenting emails into: \textit{body} (typically comprising $\sim\!\!80\%$ of the lines), \textit{header}, \textit{signoff}, \textit{signature} and \textit{greetings}.

\citet{BevendorffKPS20} introduced the Gmane corpus for email zoning\footnote{\url{https://github.com/webis-de/acl20-crawling-mailing-lists}}. Even though the corpus is composed of 31 languages,
the annotated emails are mostly in English, and their test set only contains a residual number of non-English emails (38 emails covering 13 different languages), which is insufficient for a consistent multilingual evaluation. 
Due to the richness of the Gmane conversations on technical topics, \citet{BevendorffKPS20} developed a more fine grained classification schema, considering the segmentation of blocks of code, log data and technical data. Whilst also preserving most of the common zones introduced in previous works, they ended up with a total of 15 zones:  
\textit{closing}, \textit{inline headers}, \textit{log data}, \textit{MUA signature}, \textit{paragraph}, \textit{patch}, \textit{personal signature}, \textit{quotation}, \textit{quotation marker}, \textit{raw code}, \textit{salutation}, \textit{section heading}, \textit{tabular},
 \textit{technical}, \textit{visual separator}.
Following the same zone taxonomy they also released a set of 300 English emails from the Enron database dump. %annotated with their zoning ontology.  
In both Enron and Gmane emails, the majority of the email segments belong to the \textit{paragraph} and \textit{quotation} zones. This being said, Gmane has much more lines of \textit{quotation} than \textit{paragraph}, while Enron is the other way around. 

Overall, email zoning corpora show a great variability of zone taxonomies and most works have introduced new zones to face the nature of each email source or downstream task. The Enron database dump has 
been
the most used source to retrieve emails 
to build new corpus.
On the other hand, the recent Gmane raw dump of emails is multilingual and it contains various functional zones, which opens the door to
new challenges in email zoning and multilingual methodologies.

\subsection{Cleverly Zoning Corpus}

\begin{table}[h!]
\centering
\begin{tabular}{lrrr}
\hline
\multicolumn{1}{l}{} & \textbf{pt}    & \textbf{es}   & \textbf{fr}   \\ \specialrule{1.5pt}{1pt}{1pt} 
\# zones              & 15    & 14   & 14   \\
\# emails             & 210   & 200  & 215  \\
\# lines              & 12366 & 9824 & 6958 \\
 \hline
\# lines / email         & 58.9    & 49.1   & 32.4   \\
%\# zones  /email           & 8.57  & 6.52 & 5.93 \\
\# zones  /email           & 8.6  & 6.5 & 5.9 \\
\# unique zones / email        & 5.8   & 5.1 & 4.9 \\ \hline
\end{tabular}
\caption{Some statistics of the Cleverly zoning corpus. % that were used in the experiments
}
\label{tab:annot/mcorpus-stats}
\end{table}

This section presents Cleverly zoning corpus, the first multilingual email zoning corpus.
To create the corpus,
%To build a multilingual email corpus, 
we searched the Gmane raw corpus \cite{BevendorffKPS20} for Portuguese (pt), Spanish (es) and French (fr) emails. Then, following the classification schema proposed by \citet{BevendorffKPS20}, we produced a total of 625 annotated emails. 

Table \ref{tab:annot/mcorpus-stats} compiles a brief description of the email statistics for each of the languages. While French is the language with more emails, Portuguese and Spanish emails tend to be longer, resulting in a greater amount of lines and an overall higher number of zones per email. 
The distribution of zones is similar between the three languages, as detailed in Table \ref{tab:cleverly_zone_stats}.

%\textit{quotation} ($52.2\%$ of the all the corpus' lines), \textit{paragraph} ($19.8\%$), \textit{MUA signature} ($8.8\%$), \textit{personal signature} ($3.7\%$), \textit{visual separator} ($2.7\%$), \textit{quotation marker} ($2.7\%$), \textit{closing} ($2.5\%$), \textit{raw code} ($2.0\%$), \textit{inline headers} ($1.9\%$), \textit{salutation} ($1.0\%$), \textit{tabular} ($0.4\%$), \textit{technical} ($0.3\%$), \textit{patch} ($0.2\%$) and \textit{section heading} ($0.1\%$). Spanish and French tickets do not contain any section heading lines.

\begin{table}[h!]
\centering
\label{tab:cleverly_zone_stats}
\begin{tabular}{lrrr} 
\hline
\textbf{Zone}  & \multicolumn{1}{c}{\textbf{pt (\%)} } & \multicolumn{1}{c}{\textbf{es (\%)} } & \multicolumn{1}{c}{\textbf{fr (\%)} }  \\ 
\hline
Quotation      & 52.43                     & 59.02                     & 46.20                     \\
Paragraph      & 16.33                     & 17.36                     & 27.61                     \\
MUA Sig.       & 12.04                     & 3.84                       & 9.04                       \\
Personal Sig.  & 3.93                      & 4.47                      & 2.00                       \\
Visual Sep.    & 2.94                       & 2.29                       & 2,60                       \\
Quot. Mark.    & 2.72                       & 1.54                       & 2.10                       \\
Closing        & 2.63                       & 2.00                       & 3.73                      \\
Log Data       & 1.04                       & 3.79                       & 1.82                       \\
Raw Code       & 1.28                      & 2.45                       & 2.07                       \\
Inl. Head.     & 2.96                     & 0,82                       & 1.33                       \\
Salutation     & 0,96                       & 0.81                     & 1,35                       \\
Tabular        & 0.32                       & 0.42                       & 0.27                       \\
Technical      & 0.30                       & 1.00                 & 0.38                       \\
Patch          & 0.02                         & 0.20                         & 0.02                          \\
Sec. Head.     & 0.15                      & 0.04                           & 0.03                            \\
\hline
\end{tabular}
\caption{Distribution, for each language, of the number of lines per zone in the Cleverly zoning corpus. The distributions were obtained by averaging statistics from both annotators.}
\label{tab:cleverly_zone_stats}
\end{table}

\begin{table}[h!]
\centering
\label{tab:english_2_5_zones}
\begin{tabular}{llll} 
\hline
\multicolumn{1}{c}{\textbf{measure}} & \multicolumn{1}{c}{\textbf{pt}} & \multicolumn{1}{c}{\textbf{es}} & \multicolumn{1}{c}{\textbf{fr}}
\\ 
\hline
accuracy         &  0.93 & 0.92 & 0.96  \\
$F_1$~A$_1$A$_2$  & 0.93   & 0.92   & 0.96                \\
$F_1$~A$_2$A$_1$    & 0.94  &  0.92      & 0.96    \\
\textit{$k$}    & 0.90      &  0.87      & 0.94  \\
\hline
\end{tabular}
\caption{Inter-annotator agreement for each language in the Cleverly zoning corpus, using Cohen's kappa ($k$), accuracy and $F_1$ between annotators A$_1$ and A$_2$.}
\label{tab:annotator}
\end{table}

The annotation was carried out by two annotators. The first annotator was a native Portuguese speaker and the second annotator a native Spanish speaker, both with academical background in French and fluent in the third language. Each email was annotated by both annotators using the tagtog\footnote{\url{https://www.tagtog.net}} annotation tool.

Table \ref{tab:annotator} shows the inter-annotator agreement scores for each language using the Cohen's kappa coefficient ($k$) \cite{McHugh2012InterraterRT}, accuracy and $F_1$ of one annotator versus the other. 
All annotations and required information to compile the original emails are freely available at \url{https://github.com/cleverly-ai/multilingual-email-zoning}.

\section{{\sc{Okapi}} Architecture}
\label{sec:okapi-model}

We propose {\sc{Okapi}}, an email segmentation model composed of two building blocks: a multilingual sentence encoder and a segmentation module. Figure \ref{fig:Okapi-Model} shows the {\sc{Okapi}} architecture.

\subsection{Multilingual Sentence Encoder}

\begin{figure}[h]
  \centering
  \includegraphics[width=0.55\textwidth]{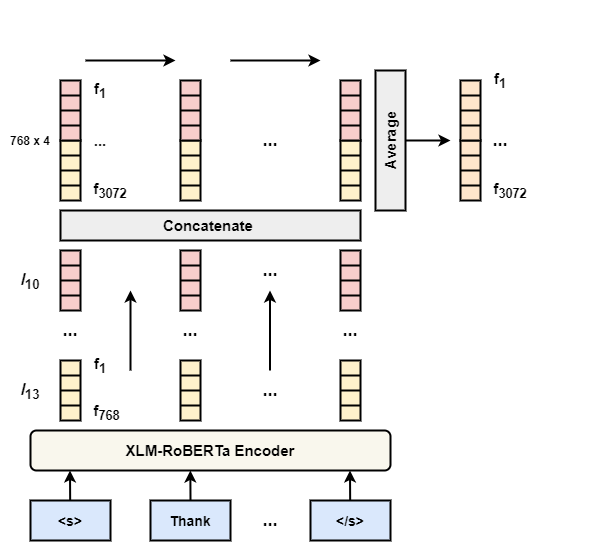}
  \caption{To derive a cross-lingual line embedding we use XLM-RoBERTa \cite{conneau2019unsupervised} to extract word-level embeddings, and then apply average pooling to the last 4 layers. This  leads to a final 3072 features embedding. }
  \label{fig:sentence-encoder}
\end{figure}

To address the multilingual nature of emails we developed a language agnostic sentence encoder that turns each email line into an embedding. 
Figure \ref{fig:sentence-encoder} %in the appendix A 
illustrates the encoding process.

Given an email line $x = \left[x_0, x_1, ..., x_n\right]$, our encoder module uses XLM-RoBERTa (base) \cite{conneau2019unsupervised} to produce an embedding $\bm{e}_{j}^{(\ell)}$ for each token $x_j$ and each layer $\ell \in \{0,1,...,13\}$. Since it has been shown that BERT-like models capture within the network layers diverse linguistic information, and, particularly, the last layers preserve most of the semantic information \cite{tenney-etal-2019-bert}, we keep, for each sentence, only the word embeddings from the last 4 layers. Lastly, as in \cite{reimers-gurevych-2019-sentence}, these word embedding are turned into a 3072 sentence embedding $\bm{s_k}$
by 
averaging the concatenation of the 4 word layer embeddings.

\subsection{Segmentation Module}

After passing each email line into the previous sentence encoder we get a cross-lingual line embedding $\bm{s_k}$. After that, we pass all line embeddings of an email 
into a Bidirectional Long Short-Term Memory (BiLSTM) \cite{Graves05}, with 1 layer and 64 hidden units, to derive compact line representations that encompass information from the entire structure of the email. Finally, as in \citet{Huang15}, we use a CRF output layer to predict the zone of each line in the document. %\bj{Reviewed}
Preliminary experiments showed that not using CRF either slightly deteriorates model performance or does not have an impact on the results.

\subsection{Training setup}
During training, XLM-RoBERTa's weights were kept frozen and only the BiLSTM and CRF layers were updated. %During the development of {\sc{Okapi}} 
We experimented BiLSTM with $16, 32, 64, 128, 256$ and $512$
hidden units and more layers, 
but in the end, having a small segmentation module, with $64$ hidden units and $1$ layer, generically yielded the best performances in % the Eron and ASF
the validation splits. 
We used a dropout layer of value 0.25 between the BiLSTM and the CRF, and the RMSprop
optimizer with a fixed learning rate of 0.001.

\section{Results and Discussion}
\label{sec:results_discussion}
In this section, we analyse both multilingual and monolingual capabilities of {\sc{Okapi}}, considering various zoning corpora and annotation schemas.

\subsection{Multilingual Email Zoning}

\begin{table}[h!]
\centering
\begin{tabular}{lccclc}
\hline
 \textbf{zone}             & \textbf{pt} &           & \textbf{es} &           & \textbf{fr} \\ \specialrule{1.5pt}{1pt}{1pt}
All    & 0.91                & \textbf{} & 0.93             & \textbf{} & 0.93            \\ \hline
Quotation     & 0.99               & \textbf{} & 0.99             & \textbf{} & 0.99            \\
Paragraph     & 0.91               & \textbf{} & 0.96             & \textbf{} & 0.92            \\
MUA Sig.      & 0.95              & \textbf{} & 0.82             & \textbf{} & 0.91  \\
Personal Sig. & 0.81                & \textbf{} & 0.87             & \textbf{} & 0.79           \\
Visual Sep. & 0.92               & \textbf{}  & 0.90             & \textbf{} & 0.96            \\
Quot. Mark. & 0.55               & \textbf{}  & 0.97             & \textbf{} & 0.97          \\
Closing & 0.59              & \textbf{}  & 0.58             & \textbf{} & 0.69            \\
Log Data      & 0.56              & \textbf{} & 0.53             & \textbf{} & 0.57           \\
Raw Code      & 0.54                & \textbf{} & 0.74             & \textbf{} & 0.84            \\ 
Inl. Head.      & 0.78                & \textbf{} & 0.77             & \textbf{} & 0.58            \\ 
Salutation      & 0.65                & \textbf{} & 0.69             & \textbf{} & 0.89            \\ 
Tabular      & 0.30                & \textbf{} & 0.00             & \textbf{} & 0.60  \\ 
Technical      & 0.67                & \textbf{} & 0.56             & \textbf{} & 0.48     \\
Patch      & 0.00                & \textbf{} & 0.00             & \textbf{} & 0.00     \\
Sec. Head.      & 0.34                & \textbf{} & 0.00             & \textbf{} & 0.00     \\\hline
\end{tabular}
\caption{Multilingual zero-shot evaluation of  {\sc{Okapi}}, using Cleverly zoning corpus. Global accuracy and recall of each zone, computed by averaging the scores regarding both annotators.}
\label{tab:multilingual}
\end{table}

We evaluate the multilingual capabilities of  {\sc{Okapi}} in a zero-shot fashion. For that, we trained the model with the %training partition of 
 Gmane English corpus released by \citet{BevendorffKPS20}, and tested it with the Cleverly multilingual corpus that we annotated for Portuguese, Spanish and French.

Table \ref{tab:multilingual} presents the performances of 
{\sc{Okapi}} in our multilingual corpus for each zone. 
Comparing with the typical performance of email zoning and the Gmane corpus (see next Tables), {\sc{Okapi}} achieves quite reasonable performances, confirming its multilingual character. 
As expected, zone recall seems to be dependent on the
total number of lines per zone.

\subsection{English Email Zoning}

\begin{table}[h!]
\centering
\begin{tabular}{lccc}
\hline
\textbf{Model} & \textbf{Zones} & \textbf{Enron} & \textbf{ASF} \\
\specialrule{1.5pt}{1pt}{1pt}
{\sc{Jangada}} & 2 & 0.88 & 0.97 \\
{\sc{Zebra}} & 2 & 0.25 & 0.18 \\
{\sc{Quagga}}  & 2 & 0.98 & 0.98 \\
\textbf{{\sc{Okapi}}} & 2 & \textbf{0.99} & \textbf{0.99} \\ \hline
{\sc{Jangada}} & 5 & 0.85 & 0.91 \\
{\sc{Zebra}} & 5 & 0.24 & 0.20 \\
{\sc{Quagga}} & 5 & 0.93 & \textbf{0.95} \\
\textbf{{\sc{Okapi}}} & 5 & \textbf{0.96} & \textbf{0.95} \\ \hline
\end{tabular}
\caption{Email zoning accuracy of various models, for the corpus of \citet{RepkeK18}.}
\label{tab:english_2_5_zones}
\end{table}

\begin{table}[h!]
\centering
\begin{tabular}{l@{\hspace*{0.25cm}}ccc}
\hline
\textbf{Model}   & \textbf{Zones} & \textbf{Gmane} & \textbf{Enron} \\
\specialrule{1.5pt}{1pt}{1pt} 
\citet{Tang2005}           & 15 & 0.80      & 0.73  \\
{\sc{Quagga}}         & 15 & 0.94          & 0.83  \\
{\sc{Chipmunck}}      & 15   & \textbf{0.96} & \textbf{0.88} \\
{\sc{Okapi}}          &  15 & \textbf{0.96} & \textbf{0.88}\\
\hline
\end{tabular}
\caption{Zoning accuracy of various models, under the 15-level zoning schema of \citet{BevendorffKPS20}. 
}
\label{tab:english_comparison_gmane_enron}
\end{table}

Resorting to the numbers %and experimental setup
reported in the literature for email zoning, we compared {\sc{Okapi}} with existing monolingual methods using various English corpora and zoning taxonomies. In particular, Table \ref{tab:english_2_5_zones} compares {\sc{Okapi}} with other zoning systems on the corpora annotated by \citet{RepkeK18} with 2 and 5 types of zones; and Table \ref{tab:english_comparison_gmane_enron} shows the results obtained with the most recent and fine-grained annotation schema with 15 zones proposed by \citet{BevendorffKPS20}. For all those combination of corpora and zoning strategies, {\sc{Okapi}} achieved competitive, and sometimes better results when compared with state-of-the-art methods for English email zoning, being simultaneously able to perform well on different languages.

\begin{table}[h!]
\begin{tabular}{l@{\hspace*{0.25cm}}c@{\hspace*{0.25cm}}c@{\hspace*{0.25cm}}c}
%\toprule
\hline
\textbf{}  & \textbf{Corpus}  & \textbf{Accuracy} & \textbf{Accuracy}\\

\textbf{Model}  & \textbf{ Train/Test}  & \textbf{ 2 zones} & \textbf{5 zones} \\ 
%\midrule
\specialrule{1.5pt}{1pt}{1pt} 
{\sc{Quagga}}            & Enron/ASF  & 0.94      & 0.86                      \\
%\textbf{Ours}  
{\sc{Okapi}} & Enron/ASF  & \textbf{0.98}             & \textbf{0.93}             \\ \midrule
{\sc{Quagga}}           & ASF/Enron   & 0.86     & 0.80             \\
%\textbf{Ours}  
{\sc{Okapi}} & ASF/Enron   & \textbf{0.97}             & \textbf{0.88}             \\ \bottomrule
\end{tabular}
\caption{Comparison between {\sc{Okapi}} and {\sc{Quagga}} for  domain adaptation, considering \citet{RepkeK18} 2 and 5 zoning schema.}
\label{tabela:gen}
\end{table}

Finally, we analyse  how {\sc{Okapi}} adapts to new domains. For that, Table \ref{tabela:gen} shows the performance of both {\sc{Okapi}} and {\sc{Quagga}} \cite{RepkeK18}, when evaluated in a different corpus then the one %the one
they were trained on. In these experiments, {\sc{Okapi}} clearly outperformed {\sc{Quagga}}, indicating a superior ability to generalize to unseen domains.

\section{Conclusion}
\label{sec:conclusion}

To overcome the English-centric email zoning literature we propose {\sc{Okapi}}. Besides having multilingual capabilities, the proposed model is competitive with existing approaches for English email zoning, and attained state-of-the-art performance in domain adaptation tasks of English email zoning. 
Futhermore, to evaluate our model and to foster future research into multilingual email zoning, we release Cleverly zoning corpus -- a corpus with 625 emails annotated in Portuguese, Spanish and French.

\section{Acknowledgments}
This project has received funding from the European Union’s Horizon 2020 research and innovation program under grant agreement No 873904.

\bibliography{anthology,referencias} %eacl2021

\begin{thebibliography}{29}
\expandafter\ifx\csname natexlab\endcsname\relax\def\natexlab#1{#1}\fi

\bibitem[{Bettenburg et~al.(2011)Bettenburg, Adams, Hassan, and
  Smidt}]{Bettenburg11}
Nicolas Bettenburg, Bram Adams, Ahmed~E. Hassan, and Michel Smidt. 2011.
\newblock \href {https://doi.org/10.1109/ICPC.2011.36} {A lightweight approach
  to uncover technical artifacts in unstructured data}.
\newblock In \emph{International Conference on Program Comprehension}, pages
  185--188, Los Alamitos, CA, USA. IEEE Computer Society.

\bibitem[{Bevendorff et~al.(2020)Bevendorff, Al~Khatib, Potthast, and
  Stein}]{BevendorffKPS20}
Janek Bevendorff, Khalid Al~Khatib, Martin Potthast, and Benno Stein. 2020.
\newblock \href {https://doi.org/10.18653/v1/2020.acl-main.108} {Crawling and
  preprocessing mailing lists at scale for dialog analysis}.
\newblock In \emph{Proceedings of the 58th Annual Meeting of the Association
  for Computational Linguistics}, pages 1151--1158, Online. Association for
  Computational Linguistics.

\bibitem[{Carvalho and Cohen(2004)}]{CarvalhoC04}
Vitor~R. Carvalho and William~W. Cohen. 2004.
\newblock \href
  {https://www.cs.cmu.edu/~vitor/papers/sigFilePaper_finalversion.pdf}
  {Learning to extract signature and reply lines from email}.
\newblock In \emph{{CEAS} - 2004 (Conference on Email and Anti-Spam)}, Mountain
  View, {CA}, {USA}.

\bibitem[{Chen et~al.(1999)Chen, Hu, and Sproat}]{Chen99}
Hao Chen, Jianying Hu, and Richard~W. Sproat. 1999.
\newblock \href {https://doi.org/10.1145/326440.326442} {Integrating
  geometrical and linguistic analysis for email signature block parsing}.
\newblock \emph{ACM Transactions on Information Systems}, 17(4):343–366.

\bibitem[{Chen et~al.(2019)Chen, Lee, Bansal, Cao, Zhang, Lu, Tsay, Wang, Dai,
  Chen, Sohn, and Wu}]{Chen19}
Mia~Xu Chen, Benjamin~N. Lee, Gagan Bansal, Yuan Cao, Shuyuan Zhang, Justin Lu,
  Jackie Tsay, Yinan Wang, Andrew~M. Dai, Zhifeng Chen, Timothy Sohn, and
  Yonghui Wu. 2019.
\newblock \href {https://doi.org/10.1145/3292500.3330723} {Gmail smart compose:
  Real-time assisted writing}.
\newblock In \emph{Proceedings of the 25th ACM SIGKDD International Conference
  on Knowledge Discovery \& Data Mining}, KDD '19, page 2287–2295, New York,
  NY, USA. Association for Computing Machinery.

\bibitem[{Cho et~al.(2014)Cho, van Merri{\"e}nboer, Gulcehre, Bahdanau,
  Bougares, Schwenk, and Bengio}]{Cho14}
Kyunghyun Cho, Bart van Merri{\"e}nboer, Caglar Gulcehre, Dzmitry Bahdanau,
  Fethi Bougares, Holger Schwenk, and Yoshua Bengio. 2014.
\newblock \href {https://doi.org/10.3115/v1/D14-1179} {Learning phrase
  representations using {RNN} encoder{--}decoder for statistical machine
  translation}.
\newblock In \emph{Proceedings of the 2014 Conference on Empirical Methods in
  Natural Language Processing ({EMNLP})}, pages 1724--1734, Doha, Qatar.
  Association for Computational Linguistics.

\bibitem[{Christidis and Losada(2019)}]{Christidis2019}
Panayotis Christidis and {\'A}lvaro~G. Losada. 2019.
\newblock \href {https://doi.org/10.3390/socsci8110306} {Email based
  institutional network analysis: Applications and risks}.
\newblock \emph{The Social Sciences}, 8(11):306.

\bibitem[{Collins(2002)}]{Collins02}
Michael Collins. 2002.
\newblock \href {https://doi.org/10.3115/1118693.1118694} {Discriminative
  training methods for hidden markov models: Theory and experiments with
  perceptron algorithms}.
\newblock In \emph{Proceedings of the ACL-02 Conference on Empirical Methods in
  Natural Language Processing - Volume 10}, EMNLP '02, page 1–8, USA.
  Association for Computational Linguistics.

\bibitem[{Conneau et~al.(2020)Conneau, Khandelwal, Goyal, Chaudhary, Wenzek,
  Guzm{\'a}n, Grave, Ott, Zettlemoyer, and Stoyanov}]{conneau2019unsupervised}
Alexis Conneau, Kartikay Khandelwal, Naman Goyal, Vishrav Chaudhary, Guillaume
  Wenzek, Francisco Guzm{\'a}n, Edouard Grave, Myle Ott, Luke Zettlemoyer, and
  Veselin Stoyanov. 2020.
\newblock \href {https://doi.org/10.18653/v1/2020.acl-main.747} {Unsupervised
  cross-lingual representation learning at scale}.
\newblock In \emph{Proceedings of the 58th Annual Meeting of the Association
  for Computational Linguistics}, pages 8440--8451, Online. Association for
  Computational Linguistics.

\bibitem[{Cortes and Vapnik(1995)}]{CortesSMV95}
Corinna Cortes and Vladimir Vapnik. 1995.
\newblock \href {https://doi.org/10.1007/BF00994018} {Support-vector networks}.
\newblock \emph{Machine Learning}, 20:273--297.

\bibitem[{Elman(1990)}]{Elman90}
Jeffrey~L. Elman. 1990.
\newblock \href {https://doi.org/https://doi.org/10.1016/0364-0213(90)90002-E}
  {Finding structure in time}.
\newblock \emph{Cognitive Science}, 14(2):179 -- 211.

\bibitem[{Estival et~al.(2007)Estival, Gaustad, Pham, Radford, and
  Hutchinson}]{Estival07authorprofiling}
Dominique Estival, Tanja Gaustad, Son~Bao Pham, Will Radford, and Ben
  Hutchinson. 2007.
\newblock Author profiling for english emails.
\newblock In \emph{Proceedings of the 10th Conference of the Pacific
  Association for Computational Linguistics}, pages 263--272.

\bibitem[{Graves and Schmidhuber(2005)}]{Graves05}
Alex Graves and J{\"{u}}rgen Schmidhuber. 2005.
\newblock \href {https://doi.org/10.1016/j.neunet.2005.06.042} {Framewise
  phoneme classification with bidirectional {LSTM} and other neural network
  architectures}.
\newblock \emph{Neural Networks}, 18(5):602--610.

\bibitem[{Huang et~al.(2015)Huang, Xu, and Yu}]{Huang15}
Zhiheng Huang, Wei Xu, and Kai Yu. 2015.
\newblock \href {http://arxiv.org/abs/1508.01991} {Bidirectional {LSTM-CRF}
  models for sequence tagging}.
\newblock \emph{CoRR}, abs/1508.01991.

\bibitem[{Kannan et~al.(2016)Kannan, Kurach, Ravi, Kaufmann, Tomkins, Miklos,
  Corrado, Lukacs, Ganea, Young, and Ramavajjala}]{45189}
Anjuli Kannan, Karol Kurach, Sujith Ravi, Tobias Kaufmann, Andrew Tomkins,
  Balint Miklos, Greg Corrado, Laszlo Lukacs, Marina Ganea, Peter Young, and
  Vivek Ramavajjala. 2016.
\newblock \href {https://doi.org/10.1145/2939672.2939801} {Smart reply:
  Automated response suggestion for email}.
\newblock In \emph{Proceedings of the 22nd ACM SIGKDD International Conference
  on Knowledge Discovery and Data Mining}, KDD '16, page 955–964, New York,
  NY, USA. Association for Computing Machinery.

\bibitem[{Klimt and Yang(2004)}]{Lang95}
Bryan Klimt and Yiming Yang. 2004.
\newblock \href {https://doi.org/10.1007/978-3-540-30115-8_22} {The enron
  corpus: A new dataset for email classification research}.
\newblock In \emph{Proceedings of the 15th European Conference on Machine
  Learning}, ECML'04, page 217–226, Berlin, Heidelberg. Springer-Verlag.

\bibitem[{Kocayusufoglu et~al.(2019)Kocayusufoglu, Sheng, Vo, Wendt, Zhao,
  Tata, and Najork}]{Kocayusufoglu19}
Furkan Kocayusufoglu, Ying Sheng, Nguyen Vo, James Wendt, Qi~Zhao, Sandeep
  Tata, and Marc Najork. 2019.
\newblock \href {https://doi.org/10.1145/3308558.3313720} {Riser: Learning
  better representations for richly structured emails}.
\newblock In \emph{The World Wide Web Conference}, WWW '19, page 886–895, New
  York, NY, USA. Association for Computing Machinery.

\bibitem[{Lafferty et~al.(2001)Lafferty, McCallum, and Pereira}]{Lafferty2001}
John~D. Lafferty, Andrew McCallum, and Fernando C.~N. Pereira. 2001.
\newblock Conditional random fields: Probabilistic models for segmenting and
  labeling sequence data.
\newblock In \emph{Proceedings of the Eighteenth International Conference on
  Machine Learning}, ICML '01, page 282–289, San Francisco, CA, USA. Morgan
  Kaufmann Publishers Inc.

\bibitem[{Lampert et~al.(2009)Lampert, Dale, and Paris}]{LampertDP09}
Andrew Lampert, Robert Dale, and C{\'e}cile Paris. 2009.
\newblock \href {https://doi.org/10.3115/1699571.1699632} {Segmenting email
  message text into zones}.
\newblock In \emph{Proceedings of the 2009 Conference on Empirical Methods in
  Natural Language Processing: A Meeting of SIGDAT, EMNLP 2009}, pages
  919--928. Association for Computational Linguistics (ACL).

\bibitem[{Lampert et~al.(2010)Lampert, Dale, and
  Paris}]{lampert-etal-2010-detecting}
Andrew Lampert, Robert Dale, and Cecile Paris. 2010.
\newblock \href {https://www.aclweb.org/anthology/N10-1142} {Detecting emails
  containing requests for action}.
\newblock In \emph{Human Language Technologies: The 2010 Annual Conference of
  the North {A}merican Chapter of the Association for Computational
  Linguistics}, pages 984--992, Los Angeles, California. Association for
  Computational Linguistics.

\bibitem[{LeCun et~al.(1989)LeCun, Boser, Denker, Henderson, Howard, Hubbard,
  and Jackel}]{LeCun89}
Yann LeCun, Bernhard~E. Boser, John~S. Denker, Donnie Henderson, Richard~E.
  Howard, Wayne~E. Hubbard, and Lawrence~D. Jackel. 1989.
\newblock \href {https://doi.org/10.1162/neco.1989.1.4.541} {Backpropagation
  applied to handwritten zip code recognition}.
\newblock \emph{Neural Computation}, 1(4):541–551.

\bibitem[{McHugh(2012)}]{McHugh2012InterraterRT}
M.~McHugh. 2012.
\newblock Interrater reliability: the kappa statistic.
\newblock \emph{Biochemia Medica}, 22:276 -- 282.

\bibitem[{Proskurnia et~al.(2017)Proskurnia, Cartright, Garcia-Pueyo, Krka,
  Wendt, Kaufmann, and Miklos}]{Proskurnia17}
Julia Proskurnia, Marc-Allen Cartright, Lluis Garcia-Pueyo, Ivo Krka, James~B.
  Wendt, Tobias Kaufmann, and Balint Miklos. 2017.
\newblock \href {https://doi.org/10.1145/3038912.3052631} {Template induction
  over unstructured email corpora}.
\newblock In \emph{Proceedings of the 26th International Conference on World
  Wide Web}, WWW '17, page 1521–1530, Republic and Canton of Geneva, CHE.
  International World Wide Web Conferences Steering Committee.

\bibitem[{Qaroush et~al.(2012)Qaroush, Khater, and Washaha}]{QaroushKW12}
Aziz Qaroush, Ismail~M. Khater, and Mahdi Washaha. 2012.
\newblock \href {https://doi.org/10.1145/2381716.2381863} {Identifying spam
  e-mail based-on statistical header features and sender behavior}.
\newblock In \emph{Proceedings of the CUBE International Information Technology
  Conference}, page 771–778, NY, USA. Association for Computing Machinery.

\bibitem[{Reimers and Gurevych(2019)}]{reimers-gurevych-2019-sentence}
Nils Reimers and Iryna Gurevych. 2019.
\newblock \href {https://doi.org/10.18653/v1/D19-1410} {Sentence-{BERT}:
  Sentence embeddings using {S}iamese {BERT}-networks}.
\newblock In \emph{Proceedings of the 2019 Conference on Empirical Methods in
  Natural Language Processing and the 9th International Joint Conference on
  Natural Language Processing (EMNLP-IJCNLP)}, pages 3982--3992, Hong Kong,
  China. Association for Computational Linguistics.

\bibitem[{Repke and Krestel(2018)}]{RepkeK18}
Tim Repke and Ralf Krestel. 2018.
\newblock Bringing back structure to free text email conversations with
  recurrent neural networks.
\newblock In \emph{European Conference on Information Retrieval}, pages
  114--126. Springer.

\bibitem[{Tang et~al.(2005)Tang, Li, Cao, and Tang}]{Tang2005}
Jie Tang, Hang Li, Yunbo Cao, and Zhaohui Tang. 2005.
\newblock \href {https://doi.org/10.1145/1081870.1081926} {Email data
  cleaning}.
\newblock In \emph{Proceedings of the Eleventh ACM SIGKDD International
  Conference on Knowledge Discovery in Data Mining}, KDD '05, page 489–498,
  New York, NY, USA. Association for Computing Machinery.

\bibitem[{Tenney et~al.(2019)Tenney, Das, and Pavlick}]{tenney-etal-2019-bert}
Ian Tenney, Dipanjan Das, and Ellie Pavlick. 2019.
\newblock \href {https://doi.org/10.18653/v1/P19-1452} {{BERT} rediscovers the
  classical {NLP} pipeline}.
\newblock In \emph{Proceedings of the 57th Annual Meeting of the Association
  for Computational Linguistics}, pages 4593--4601, Florence, Italy.
  Association for Computational Linguistics.

\bibitem[{Vaswani et~al.(2017)Vaswani, Shazeer, Parmar, Uszkoreit, Jones,
  Gomez, Kaiser, and Polosukhin}]{NIPS2017_3f5ee243}
Ashish Vaswani, Noam Shazeer, Niki Parmar, Jakob Uszkoreit, Llion Jones,
  Aidan~N Gomez, \L~ukasz Kaiser, and Illia Polosukhin. 2017.
\newblock \href
  {https://proceedings.neurips.cc/paper/2017/file/3f5ee243547dee91fbd053c1c4a845aa-Paper.pdf}
  {Attention is all you need}.
\newblock In \emph{Advances in Neural Information Processing Systems},
  volume~30, pages 5998--6008. Curran Associates, Inc.

\end{thebibliography}
\bibliographystyle{acl_natbib}

\end{document}